\documentclass[10pt,twocolumn,letterpaper]{article}
\usepackage[accsupp]{axessibility}
\usepackage{iccv}
\usepackage{times}
\usepackage{epsfig}
\usepackage{graphicx}
\usepackage{amsmath}
\usepackage{amssymb}
\usepackage{multirow}
\usepackage{booktabs}
\usepackage{bbm}

\usepackage{pifont}
\newcommand{\cmark}{\ding{51}}%
\usepackage[pagebackref=true,breaklinks=true,colorlinks,bookmarks=false]{hyperref}
\usepackage{xcolor}
\PassOptionsToPackage{dvipsnames}{xcolor}
\RequirePackage{xcolor} 
\definecolor{halfgray}{gray}{0.55} 
\definecolor{webgreen}{rgb}{0,.5,0}
\definecolor{webbrown}{rgb}{.6,0,0}
\usepackage{subfigure}

\usepackage[breaklinks=true,bookmarks=false]{hyperref}

\iccvfinalcopy 

\def\iccvPaperID{****} 

\ificcvfinal\fi

\begin{document}

\title{Action Sensitivity Learning for Temporal Action Localization}


\author{Jiayi Shao$^{1,}$\thanks{~This work was done during the first author's internship in Alibaba.},  ~ Xiaohan Wang$^{1}$, Ruijie Quan$^{1}$, Junjun Zheng$^{2}$, Jiang Yang$^{2}$, Yi Yang$^{1,}$\thanks{~Corresponding Author.} \\
$^1$ReLER Lab, CCAI, Zhejiang University, $^2$Alibaba Group \\
{\tt\small{shaojiayi1@zju.edu.cn, wxh1996111@gmail.com, quanruij@hotmail.com}}, \\ {\tt\small{\{fangcheng.zjj,yangjiang.yj\}@alibaba-inc.com, yangyics@zju.edu.cn}}
 }

\maketitle
\ificcvfinal\thispagestyle{empty}\fi

\begin{abstract}
Temporal action localization (TAL), which involves recognizing and locating action instances, is a challenging task in video understanding.
Most existing approaches directly predict action classes and regress offsets to boundaries, while overlooking the discrepant importance of each frame.
In this paper, we propose an Action Sensitivity Learning framework (ASL) to tackle this task, which aims to assess the value of each frame and then leverage the generated action sensitivity to recalibrate the training procedure. We first introduce a lightweight Action Sensitivity Evaluator to learn the action sensitivity at the class level and instance level, respectively. The outputs of the two branches are combined to reweight the gradient of the two sub-tasks. Moreover, based on the action sensitivity of each frame, we design an Action Sensitive Contrastive Loss to enhance features, where the action-aware frames are sampled as positive pairs to push away the action-irrelevant frames. The extensive studies on various action localization benchmarks (i.e., MultiThumos, Charades, Ego4D-Moment Queries v1.0, Epic-Kitchens 100, Thumos14 and ActivityNet1.3) show that ASL surpasses the state-of-the-art in terms of average-mAP under multiple types of scenarios, e.g., single-labeled, densely-labeled and egocentric. 

\end{abstract}

\vspace{-0.5em}
\section{Introduction}
With an increasing number of videos appearing online, video understanding has become a prominent research topic in computer vision. Temporal action localization (TAL), which aims to temporally locate and recognize human actions with a set of categories in a video clip,  is a challenging yet fundamental task in this area, owing to its various applications such as sports highlighting, human action analysis and security monitoring~\cite{app1,app2,app3,app4,app5}.

\begin{figure}[t]
\centering
\includegraphics[height=0.46\linewidth, width=1.0\linewidth]{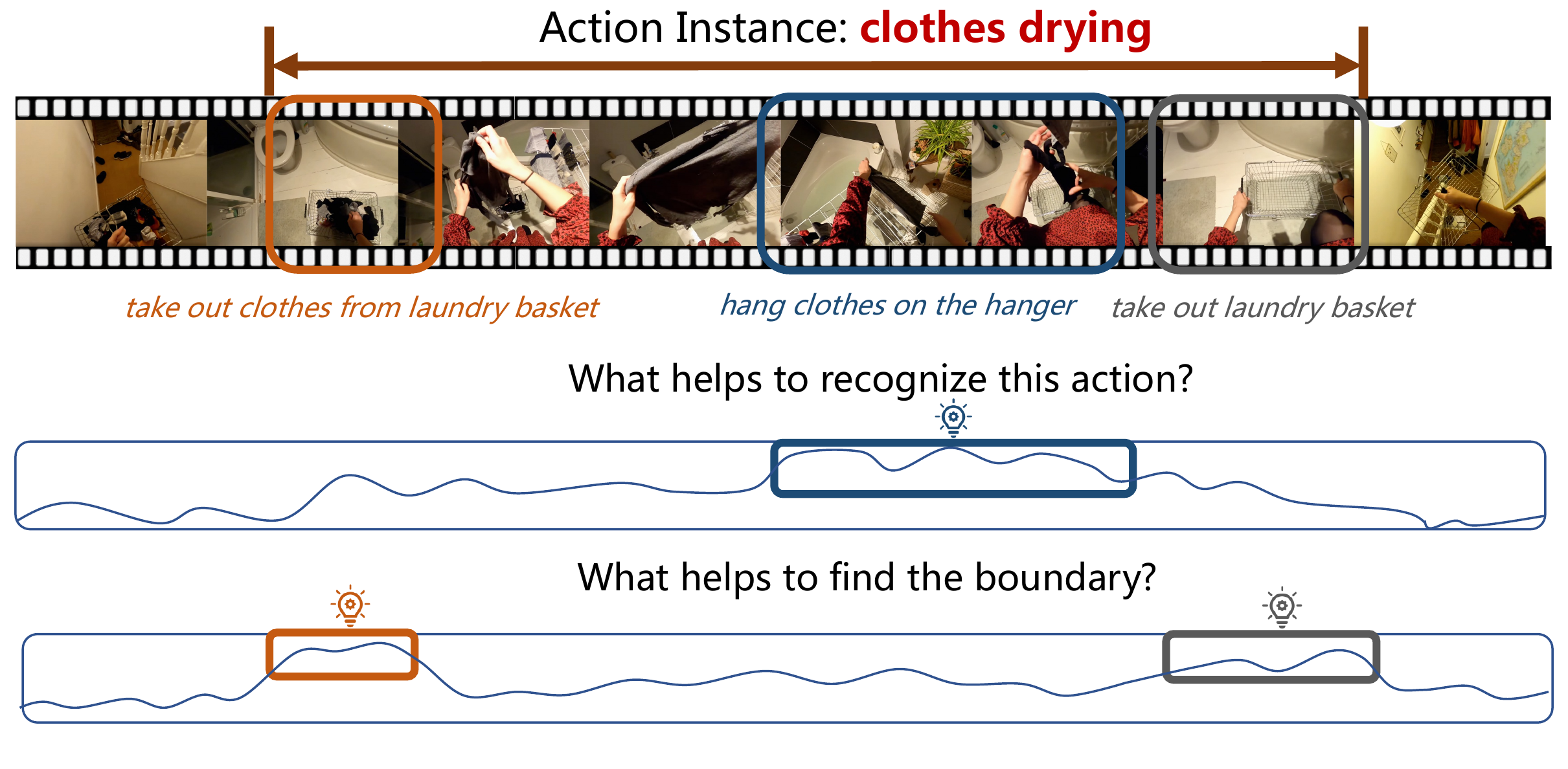}
\caption{The motivation of our method. We show the action instance of \textbf{clothes drying} and depict the possible importance of each frame to recognizing the action category and locating action boundaries. Each frame's importance is different. }
\label{fig:intro}
\vspace{-1em}
\end{figure}

\par We have recently witnessed significant progress in TAL, where most methods can be mainly divided into two parts: 1) Two-stage approaches~\cite{PGCN2019ICCV, contextloc} tackle this task accompanied by the generation of class-agnostic action proposals and then perform classification and proposal boundaries refinement in proposal-level; 2) One-stage approaches~\cite{vsgn,xu2020gtad, ssd} simultaneously recognize and localize action instances in a single-shot manner. Typical methods~\cite{zhang2022actionformer,afsd} of this type predict categories as well as locate corresponding temporal boundaries in frame-level, achieving stronger TAL results currently.~In training, they classify every frame as one action category or background and regress the boundaries of frames inside ground-truth action segments. However, these works treat each frame within action segments equally in training, leading to sub-optimal performance.
\label{sec:intro}
\par  When humans intend to locate action instances, the discrepant information of each frame is referred to.  For the instance of action:~\textbf{clothes drying}, as depicted in Fig~\ref{fig:intro},  frames in the purple box promote recognizing \textbf{clothes drying} most as they describe the intrinsic sub-action: \textit{hang clothes on the hanger}. Analogously, frames in red and gray boxes depict \textit{take out clothes from laundry basket} and \textit{lift laundry basket}, which are more informative to locate precise start and end time respectively.
In a word, each frame's contribution is quite different,  due to intrinsic patterns of actions, as well as existing transitional or blurred frames.
\par \textit{Can we discover informative frames for classifying and localizing respectively?} To this end, we first introduce a concept --- \textbf{Action Sensitivity}, to measure the frame's importance. It is disentangled into two parts: \textbf{action sensitivity to classification} sub-task and \textbf{action sensitivity to localization} sub-task. For one sub-task, the higher action sensitivity each frame has, the more important it will be for this sub-task. With this concept, intuitively, more attention should be paid to action sensitive frames in training.
\par Therefore in this paper, we propose a lightweight Action Sensitivity Evaluator (ASE) for each sub-task to better exploit frame-level information. Essentially, for a specific sub-task, ASE learns the action sensitivity of each frame from two perspectives: class-level and instance-level. The class-level perspective is to model the coarse action sensitivity distribution of each action category and is achieved by incorporating gaussian weights. The instance-level perspective is complementary to class-level modeling and is supervised in a prediction-aware manner. Then the training weights of each frame are dynamically adjusted depending on their action sensitivity, making it more reasonable and effective for model training.
\par With the proposed ASE, we build our novel \underline{A}ction \underline{S}ensitivity \underline{L}earning framework dubbed \textbf{ASL} to tackle temporal action localization task (TAL) effectively.  Moreover, to furthermore enhance the features and improve the discrimination between actions and backgrounds, we design a novel Action Sensitive Contrastive Loss (ASCL) based on ASE. It is implemented by elaborately generating various types of action-related and action-irrelevant features and performing contrasting between them, which brings multiple merits for TAL.
\par By conducting extensive experiments on 6 datasets and detailed ablation studies, we demonstrate ASL is able to classify and localize action instances better. In a nutshell, our main contributions can be summarized as follows:
\begin{itemize}
\item[$\bullet$] We propose a novel framework with an Action Sensitivity Evaluator component to boost training,  by discovering action sensitive frames to specific sub-tasks, which is modeled from class level and instance level.
\vspace{-0.2em}
\item[$\bullet$] We design an Action Sensitive Contrastive Loss to do feature enhancement and to increase the discrimination between actions and backgrounds.
\vspace{-0.2em}
\item[$\bullet$] We verify ASL on various action localization datasets of multiple types: i) densely-labeled (i.e., MultiThumos~\cite{multithumos} and Charades~\cite{charades}). ii) egocentric (Ego4d-Moment Queries v1.0~\cite{grauman2022ego4d} and Epic-Kitchens 100~\cite{Damen2018EPICKITCHENS}). iii) nearly single-labeled (Thumos14~\cite{thumos} and ActivityNet1.3~\cite{caba2015activitynet}), and achieve superior results.
\end{itemize}

\section{Related Works}
\par \textbf{Temporal Action Localization.} Temporal action localization is a long-standing research topic. Contemporary approaches mostly fall into two categories, i.e. two-stage and one-stage paradigms. Previous two-stage methods usually focused on action proposal generation~\cite{lin2019bmn,BSN2018arXiv,su2021bsn++, rtdnet,wang2022rcl}. Others have integrated action proposal, calibrated backbone, classification and boundary regression or refinement modules into one single model~\cite{shou2016temporalmscnn,rc3d,shou2017cdc,SSN2017ICCV}.  Recent efforts have investigated the proposal relations~\cite{PGCN2019ICCV, contextloc, relation},   utilized graph modeling~\cite{xu2020gtad,PGCN2019ICCV}, or designed fine-grained temporal representation~\cite{tcanet,sridhar2021class}. One-stage approaches usually perform frame-level or segment-level classification and directly localization or merging segments~\cite{shou2017cdc, bottomup, ssd}. \cite{vsgn,gtan} process the video with the assistance of pre-defined anchors or learned proposals, while others utilize existing information and are totally anchor-free ~\cite{afsd,zhang2022actionformer,segtad}. Currently, some works introduce pretrain-finetune to TAL task~\cite{boundarysensitivepretrain,xumengmenglow} or attempt to train the model in an efficient end-to-end manner~\cite{tadtr,cheng2022tallformer,etad}. Others focused on densely-labeled setting~\cite{mlad,dai2021pdan,dai2022mstct,kahatapitiya2021coarsefine,tanpointtad,dai2021ctrn}. With the success of DETR~\cite{detr} in object detection, query-based methods have also been proposed~\cite{shi2022react,rtdnet,tanpointtad, tadtr}. Our method falls into the one-stage TAL paradigm and performs frame-level classification and localization.  Notably, \cite{tgm,gtan} incorporate Gaussian kernels to improve receptive fields and optimize the temporal scale of action proposals, \cite{kahatapitiya2021coarsefine} use fixed gaussian-like weights to fuse the coarse and fine stage. We also utilize gaussian weights as one part of ASE, but it differs in that: 
i) Our gaussian-like weights in ASE serve as modeling class-level action sensitivity and to boost effective training, while~\cite{kahatapitiya2021coarsefine,tgm,gtan} use it only to better encode the videos. 
ii) Our learned gaussian weights describe frames' contributions to each sub-task and can be easily visualized, whereas the semantic meaning of gaussian weights in~\cite{kahatapitiya2021coarsefine,tgm,gtan} is unclear.
iii) Our gaussian-like weights are totally learnable, category-aware and disentangled to different sub-tasks.
\par \textbf{One-stage Object Detection.} Analogous to TAL task, the object detection task shares a few similarities. As a counterpart in object detection, the one-stage paradigm has surged recently. Some works remain anchor-based~\cite{lin2017focal}, while others are anchor-free, utilizing a feature pyramid network~\cite{fpn,tian2019fcos} and improved label-assign strategies~\cite{atss,zhu2020autoassign,zhu2019feature,ohem}. Moreover, some works define key points in different ways (e.g. corner~\cite{law2018cornernet}, center~\cite{duan2019centernet,tian2019fcos} or learned points~\cite{yang2019reppoints}). These methods bring some inspiration to design a better TAL framework. Some methods~\cite{feng2021tood,li2020generalizedfocal1,li2021generalizedfocal2} aim to tackle the misalignment between classification and localization. But i) we mainly focus on the discrepant information of frames. ii) Misalignment of two sub-tasks (i.e., classification and localization) is only the second issue and we alleviate it by a novel contrastive loss which differs from these works.

\begin{figure*}[htbp]
\centering
\includegraphics[height=0.324\linewidth, width=1.0\linewidth]{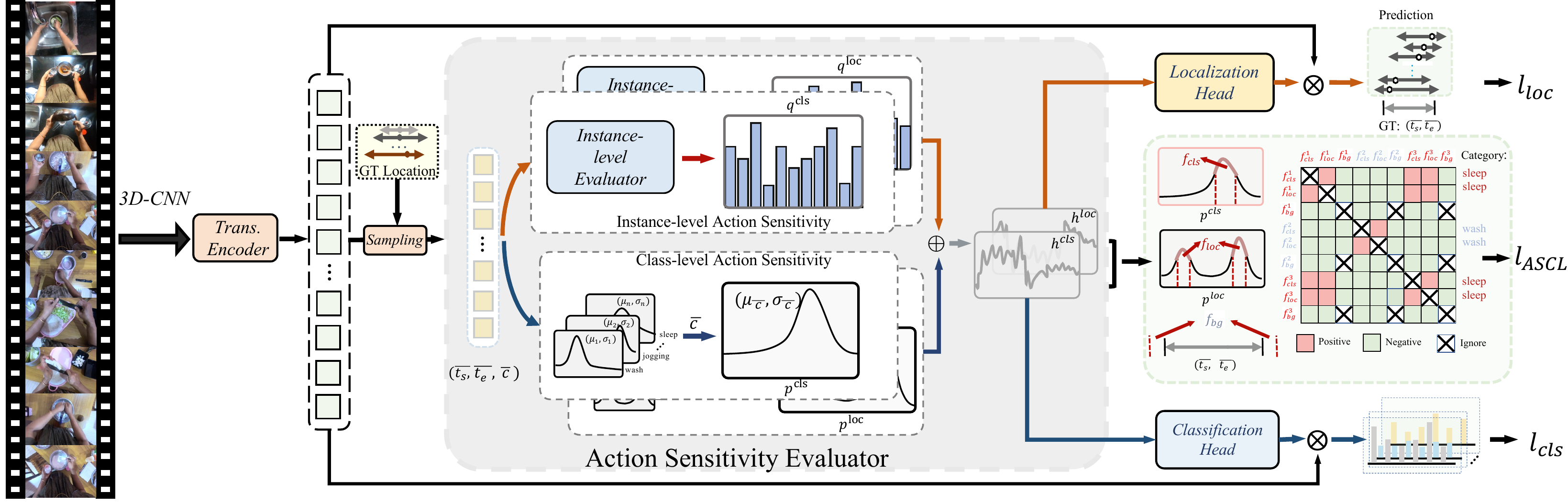}
\caption{\textbf{The overview of ASL}. Given a video clip, we first leverage a pre-trained 3D-CNN to extract the video feature and then utilize a Transformer Encoder to encode feature. We then use ground-truth location sampling to sample all ground-truth segments and feed these into Action Sensitivity Evaluator. In this module, we model sub-task-specific action sensitivity of each frame from class level and instance-level. The former is learned by incorporating learnable gaussian-like weights and the latter is learned with an instance-level evaluator. Then each frame's weight in training is adjusted based on action sensitivity. Moreover, we propose an Action Sensitive Contrastive Loss to better enhance the feature and alleviate misalignment problems.}
\label{fig:main}
\vspace{-0.75em}
\end{figure*}

\par \textbf{Contrastive Learning.} Contrastive learning \cite{contra1,contra2,contra3} is an unsupervised learning objective that aims to bring similar examples closer together in feature space while pushing dissimilar examples apart. NCE~\cite{nce} and Info-NCE~\cite{infonce} are two typical methods that mine data features by distinguishing between data and noise or negative samples. Info-NCE-based contrastive learning has been used in methods of different tasks, such as~\cite{wang2022align,reler,Wang_2021_CVPR} in cross-modality retrieval and~\cite{moco,videomoco} in unsupervised learning.  In TAL, ~\cite{afsd} leverages ranking loss to boost discrimination between foreground and background while ~\cite{shi2022react} contrasts different actions with a global representation of action segments. But we design a new contrastive loss both across different types of actions and between actions and backgrounds. Moreover, compared to~\cite{shou2018autoloc} which also contrasts between actions and backgrounds, our proposed contrastive loss contrasts more between i)same and different action classes, ii)sensitive frames of localization and classification to mitigate the misalignment of sub-tasks. Details will be discussed in~\ref{sec:ascl}.

\vspace{1.0em}
\section{Method}
\textbf{Problem Formulation.}      
The task of temporal action localization (TAL) is to predict a set of action instances $\{(t_m^s,t_m^e,c_m)\}_{m=1}^M$, given a video clip, where $M$ is the number of predicted action instances, $t_m^s$,$t_m^e$,$c_m$ are the start, end timestamp and action category of the $m$-th predicted action instance. ASL is built on an anchor-free representation that classifies each frame as one action category or background, as well as regresses the distances from this frame to the start time and end time. 

\textbf{Overview.} 
The overall architecture of ASL is shown in Fig~\ref{fig:main}. 
ASL is composed of four parts: video feature extractor, feature encoder, action sensitivity evaluator, and two sub-task heads. 
Concretely, given a video clip, we first extract the video feature using a pre-trained 3D-CNN model. Then we exert a feature encoder involving a pyramid network to better represent the temporal features at multiple levels. We propose an action sensitivity evaluator module to access the action sensitivity of frames to a specific sub-task. The pyramid features combined with frames' action sensitivity are further processed by sub-task heads to generate predictions. We now describe the details of ASL.

\subsection{Feature Encoder}
\par With the success of~\cite{zhang2022actionformer,afsd}, ASL utilizes a Transformer encoder and feature pyramid network to encode feature sequences into a multiscale representation. To enhance features, in Transformer encoder we design a new attention mechanism that operates temporal attention and channel attention parallelly and then fuses these two outputs.
\par For normal temporal attention that is performed in the temporal dimension, input features generate query, key and value tensors $(Q_t,K_t,V_t) \in \mathbb{R}^{T \times D}$, where $T$ is the number of frames, $D$ is the embedding dimension, then the output is calculated: 
\begin{equation} \small
    f_\text{ta}^{'} = \text{softmax}(\frac{Q_tK_t^T}{\sqrt{D}}) V_t
\end{equation}
\par For channel attention that is conducted in the channel dimension, input features generate query, key and value tensors $(Q_d,K_d,V_d) \in \mathbb{R}^{D \times T}$, where $D$ is the number of channels. Then the output is calculated:
\begin{equation} \small
    f_\text{ca}^{'} = \text{softmax}(\frac{Q_d K_d^T}{\sqrt{T}}) V_d
\end{equation}
\par Above two outputs are then added with a coefficient $\theta$: $f^{'} = (1-\theta)f_\text{ta}^{'} + \theta f_\text{ca}^{'T}$. Then it is processed by layer normalization and feedforward network to obtain the encoded video representation $f \in \mathbb{R}^{T\times D}$.

\subsection{Action Sensitivity Evaluator}
\par As discussed in~\ref{sec:intro}, not all frames inside ground-truth segments contribute equally to the sub-task (i.e., localization or classification). Thus we designed an Action Sensitivity Evaluator (ASE) module, the core idea of which is to determine the sub-task-specific action sensitivity of each frame and help the model pay more attention to those valuable frames. Besides, this module is lightweight, leading to efficient and effective training.
\par \textbf{Decoupling to two levels.} Digging into action instances, a key observation is that actions of a particular category often share a similar pattern, but they appear slightly different in diverse scenarios or under different behavior agents. For example,  action instances of category:\textbf{wash vegetables} inherently contain sub-actions: \textit{turn the tap on, take vegetables, wash, turn the tap off},  where frames depicting \textit{washing} are more sensitive to classification,  frames depicting \textit{turning the tap on} and \textit{turning the tap off} are more sensitive to localization. But the respective duration or proportion of these sub-actions are dependent on the scenes and context of each action instance, thus making sensitive frames a little different. This motivates us that the action sensitivity of every frame should be decoupled into class-level and instance-level modeling and then recombined from these two parts. 

\textbf{Disentangling to two sub-tasks.} Here sub-tasks mean classification and localization. Intuitively action sensitivity for classification needs to be modeled as sensitive frames for classification is not easily determined. Actually, action sensitivity modeling for localization is also necessary. Though the boundaries of action segments are defined already, sensitive frames are not necessarily at the start or the end of an action since i) action boundaries are often unclear, ii) each frame of sub-actions around boundaries also has different semantics. Therefore, action sensitivity modeling should be disentangled for two sub-tasks respectively (i.e., classification and localization). 

\par Formally, for a given ground-truth $\mathcal{G}\!=\!\{\bar t^s, \bar t^e, \bar c\}$, each indicating the start time, end time and category of one action, we denote $N_f$ as the number of frames within this action, $N_c$ as the number of all pre-defined action categories. Our goal is to model the class-level action sensitivity $p$ (disentangled into $p^{cls},p^{loc}$ to classification and localization respectively), instance-level action sensitivity $q$ (disentagled to $q^{cls}, q^{loc}$).  Then we delve into details of action sensitivity learning.

\textbf{Class-level Modeling.}  
  \ Class-level sensitivity poses a fundamental prior for action sensitivity learning. 
  Two key observations are that: i) video frames are often consecutive. ii) there often exist keyframes that have a peek value of sensitivity among all frames. 
  In this case, we incorporate gaussian-like weights with learnable parameters $\mu,\sigma \in \mathbb{R}^{N_c}$ to model class-level action sensitivity $p$.
\par For classification sub-task, we model corresponding action sensitivity  $p_{i}^{cls}$ for the $i$-th frame:
\begin{equation} \small
      p^{cls}_i = \exp\{-\frac{(d(i)-\mu_{c})^2}{2\sigma_{c}^2}\}
\end{equation}
where $d(i)$ is the distance from the current $i$-th frame to the central frame of the ground-truth segment which is normalized by $N_f$. In this case, $d(i)\! \in \![-0.5, 0.5]$, when $i\!=\!1$ (i.e., start frame), $d(i)\!=\! -0.5$, when $i\!=\!N_f$ (i.e., end frame), $d(i)\!=\!0.5$.
Learnable parameters $\mu_{c},\sigma_{c}$ denote mean and variance of each category $c$'s action sensitivity distribution.
\par For localization sub-task, different frames are sensitive to locating start time and end time. Therefore action sensitivity $p^{loc}$ is the combination of two parts.
We explicitly allocate one gaussian-like weights $p^{sot}$ to model the start time locating sensitivity and another $p^{eot}$ to model the end time locating sensitivity. $ p^{loc} $ is calculated: 
\begin{equation} \small
      p^{loc}_i = \underbrace{\exp\{-\frac{(d(i)-\mu_{c,1})^2}{2\sigma_{c,1}}\}}_{p^{sot}_i} + \underbrace{\exp\{-\frac{(d(i)-\mu_{c,2})^2}{2\sigma_{c,2}}\}}_{p^{eot}_i}
\end{equation}
\par In this way, class-level action sensitivity $p^{cls},p^{loc} \!\in \!\mathbb{R}^{N_f \times N_c}$ of all categories are learned with the optimization of model training.
In addition, the initialization of $\mu_c$ and $\sigma_c$ counts as there exists prior knowledge~\cite{zhang2022actionformer,tian2019fcos} according to different sub-tasks. For classification sub-task, near-center frames are more sensitive. Thus we initialize $\mu_c$ as 0. For localization sub-task, near-start and near-end frames are more sensitive. Thus we initialize $\mu_1$ as -0.5 and $\mu_2$ as 0.5.
For all $\sigma$, we initialize as 1.

\textbf{Instance-level Modeling.} 
Intuitively, a Gaussian can only give a single peak, and thus class-level action sensitivity learning may not discover all sensitive frames. To this end, we introduce instance-level modeling which is complementary and aims to capture additional important frames that haven’t been discovered by class-level modeling. 

In the instance-level modeling, as more information about frame contexts of each instance is referred to, we obtain instance-level action sensitivity $q \in \mathbb{R}^{N_f}$ using an instance-level evaluator operated directly on each frame, composed of 1D temporal convolution network which aims to encode temporal contexts better, a fully connected layer and a Sigmoid activation function. We denote $\Phi^{cls}$ and $\Phi^{loc}$ as two sub-task specific instance-level evaluator, then $q^{cls}$ and $q^{loc}$ are computed:  
\begin{equation} \small
\left\{
\begin{aligned}
 & q_i^{cls} = \Phi^{cls}(f_i)\\
& q_i^{loc}  = \Phi^{loc}(f_i)
\end{aligned}
\right.
\end{equation}

Unlike class-level modeling that contains some prior knowledge, instance-level sensitivity $q$ is hard to learn in an unsupervised manner. Intuitively, from the instance level a sensitive frame implies that it can result in fine predictions. Hence we utilize the quality $\{\bar Q_i\}_{i=1}^{N_f}$ of each frame's prediction to supervise the learning of $q$. 
For localization, The higher tIoU indicates a higher degree of overlap between two segments. Thus tIoU between the predicted segment and the ground-truth segment can measure the quality of prediction.
For classification, the probability of the ground-truth category can serve as the quality of prediction.
Therefore, quality  $\bar Q^{cls}$ and $\bar Q^{loc}$ are defined as: 
\begin{equation} \small
\left\{
\begin{aligned}
 \bar Q_i^{cls} &= \varphi (\text{s}_i[\bar c)]) \\
\bar Q_i^{loc} &= \text{tIoU}(\Delta_i, \bar \Delta)
\end{aligned}
\right.
\end{equation}
where $s$ denotes the classification logits, $\Delta_i$ is the predicted segment $(t^s, t^e)$ of the $i$-th frame, $\bar \Delta$ is the corresponding ground-truth segment, $\varphi(\cdot)$ is Sigmoid function. We use MSE loss to supervise the calculation of $q$. For $q^{cls}$, optimization objective is formed as~\ref{equ:ils}. Optimization of $q^{loc}$ is in a similar way.
\begin{equation} \small \label{equ:ils}
\mathcal{L}_{s} = \text{MSE} (q^{cls}, \bar Q^{cls}) 
\end{equation}

\textbf{Optimization with Action Sensitivity.} In this way, combining class-level and instance-level,  we obtain the final action sensitivity $h(\bar c) \! \in \! \mathbb{R}^{N_f}$ (disentangled to classification and localization sub-task: $h(\bar c) \rightarrow \{h^{cls}(\bar c),h^{loc}(\bar c)\}$) for the ground-truth $\mathcal{G}\!=\!\{\bar t^s, \bar t^e, \bar c\}$:
\begin{equation} \small
\left\{
\begin{aligned}
 & h^{cls}(\bar c) = p^{cls} \mathbbm{1}[\bar c] + q^{cls} \\
& h^{loc}(\bar c) = p^{loc} \mathbbm{1}[\bar c] + q^{loc}
\end{aligned}
\right.
\end{equation}
where $\mathbbm{1}[\bar c] \!\in\! \mathbb{R}^{N_c}$ denotes the one-hot vector of $\bar c$. Action sensitivity $h$ is further used in training. For classification sub-task,  we use a focal loss~\cite{lin2017focal} to classify each frame, combined with classification action sensitivity $h^{cls}$:
\begin{equation} \small \label{equ:lcls}
\mathcal{L}_{cls}= \frac{1}{N_{pos}} \sum_i (\mathbbm{1}_{in_i} h^{cls}_i(\bar c_i) \mathcal{L}_{\text{focal}_i} 
    + \mathbbm{1}_{bg_i} \mathcal{L}_{\text{focal}_i})
\end{equation}
where $\mathbbm{1}_{in_i}, \mathbbm{1}_{bg_i}$ are indicators that denote if the $i$-th frame is within one ground-truth action or if is background, $N_{pos}$ is the number of frames within action segments, $\bar c_i$ denotes the action category of the $i$-th frame.

For localization sub-task, we use a DIoU loss~\cite{zheng2020diou} performed on frames within any ground-truth action instance, to regress offsets from current frames to boundaries, combined with localization action sensitivity $h^{loc}$:
\begin{equation} \small \label{equ:lloc}
\mathcal{L}_{loc}= \frac{1}{N_{pos}} \sum_i (\mathbbm{1}_{in_i} h^{loc}_i(\bar c_i) \mathcal{L}_{\text{DIoU}_i})
\end{equation}

\subsection{Action Sensitive Contrastive Loss}
\label{sec:ascl} 
\par Now with ASE, each frame is equipped with action sensitivity and valuable frames to specific sub-tasks will be discovered. We further boost the training from the perspective of feature enhancement. 
Delve into feature representation, three shortcomings may hinder the performance: 
i) classification sensitive and localization sensitive frames are quite different, resulting in the misalignment of these two sub-tasks.  
ii) features in actions of different categories are not much discriminable.
iii) features within action and outside boundaries are not much distinguished yet.
\par Therefore, on the basis of ASE, we propose an Action Sensitive Contrastive Loss (ASCL) to correspondingly tackle the above issues.
Specifically, for a given video feature $\{f_t\}_{t=1}^T$ and a ground-truth action instance $\mathcal{G}\!=\!\{\bar t^s, \bar t^e, \bar c\}$, 
we generate two action-related features and one action-irrelevant feature. 
First, to generate more valuable action-related features, we aim to find sensitive frames to these sub-tasks. Thinking that ASCL contrasts action instances of different classes, where class-level discrimination is more important, we hence utilize class-level sensitivity $p$ to parse the sensitive frame ranges $T_{cls}$ for classification and $T_{loc}$ for localization.
With one ground-truth category $\bar c$, we get the most sensitive frames $a_{cls},a_{sot},a_{eot}$ for classification, start time localization, end time localization respectively. Take $a_{eot}$ as an example:
\begin{equation} \small
    a_{eot} = \underset{i}{\arg\max} (p_i^{eot} \mathbbm{1}[\bar c])
\end{equation}
$a_{cls}$ and $a_{sot}$ are obtained in a similar way. Then, centered on $a$ and extending forward and backward with a range of $\delta N_f$, where $\delta$ is the sampling length ratio, we get sensitive frame ranges $T_{cls}$ for classification and $T_{loc}$ for localization ($T_{cls}$ and $T_{loc}$ are limited inside the action instance).
Furthermore,  we utilize class-level sensitivity to compute sensitive features $f_{cls}$ for classification, $f_{loc}$ for localization:
\begin{equation}  \small
\left\{
\begin{aligned} 
f_{cls} &=  \frac{1}{T}\sum_{t} p^{cls}_t \mathbbm{1}[\bar c] f_t , \ t \in T_{cls} \\
f_{loc} &= \frac{1}{T} \sum_{t} p^{loc}_{t} \mathbbm{1}[\bar c]  f_{t}, \ t \in T_{loc} 
\end{aligned}
\right.
\end{equation}
\par Secondly, we aim to simultaneously discriminate actions and backgrounds better. Consequently we generate boundary-related background features $f_{bg}$:
\begin{equation} \small
    f_{bg} = \frac{1}{T} \sum_{t} f_t, \ t \in [\bar t^s-\delta N_f, \bar t^s]\cup [\bar t^e, \bar t^e+\delta N_f] 
\end{equation}
\par The learning objective of ASCL is based on a contrastive loss. As figure~\ref{fig:main} shows, the positive samples $\mathcal{P}$ are constructed from $f_{cls}$ and $f_{loc}$ in action instances of the same category while the negative samples $\mathcal{N}$ come from: i) $f_{cls}$ and $f_{loc}$ in action instances of different categories. ii) all background features $f_{bg}$. ASCL is computed for each batch $B$ with $N$ samples: 
\begin{equation} \small{ 
    \mathcal{L}_{\text{ASCL}} = \frac{1}{N}\sum_{B} -\log \frac{ \sum\limits_{f_x\in \mathcal{P}_{f_*}}\!\text{sim}(f_*, f_x) }{\sum\limits_{f_x \in \mathcal{P}_{f_*}}\! \text{sim}(f_*, f_x) +\! \sum\limits_{f_x \in \mathcal{N}_{f_*}}\! \text{sim}(f_*, f_x) }}
\end{equation}
\par Optimizing ASCL will be of benefits to tackle the corresponding issues above : 
i) alleviate the misalignment of two sub-tasks by pulling features of their respective sensitive frames closer.
ii) discriminate actions and backgrounds better by pushing action features of the same category closer and different categories apart, meanwhile pushing actions and backgrounds apart. Thus ASCL can enhance the feature representation and boost training furthermore.

\subsection{Training and Inference}
\label{sec:trainandinfer}
\textbf{Training.} In the training process, our final loss function is designed: 
\begin{equation} \small
    \mathcal{L}=\mathcal{L}_{cls} + \mathcal{L}_{loc} + \mathcal{L}_{s} + \lambda \mathcal{L}_{\text{ASCL}}
\end{equation}
where $\mathcal{L}_{cls}$, $\mathcal{L}_{loc}$ and $\mathcal{L}_{s}$ are discussed in equation~\ref{equ:lcls}, equation~\ref{equ:lloc} and equation~\ref{equ:ils}. $\lambda$ denotes the weight of Action Sensitive Contrastive loss.

\textbf{Inference.} At inference time, our model outputs predictions $(t^s,t^e,c)$ for every frame across all pyramids levels, where $t^s,t^e$ denote the start and end time of action, $c$ denote the predicted action category. $c$ also serves as the action confidence score. SoftNMS~\cite{softnms} is then applied on these results to suppress redundant predictions.

\section{Experiments}

\begin{table*}[ht]
 \centering 
 \small
 \caption{\label{tab:mthumosandcharades} \textbf{Results on MultiThumos and Charades}. We report detection-\textit{m}AP at different tIoU thresholds. Average \textit{m}AP in $[$0.1:0.1:0.9$]$ is reported on MultiThumos and Chrades. Best results are in \textbf{bold}. $\ddagger$ indicates results trained with stronger image augmentation~\cite{tanpointtad, tadtr}. I3D denotes using I3D~\cite{i3d} features and E2E indicates results trained in an end-to-end manner.}
 \vspace{0.5em}
 {
  \begin{tabular}{l|l|l|cccc|cccc} 
  \toprule
  \multirow{2}{*}{Model} & \multirow{2}{*}{Modality} & \multirow{2}{*}{Feature} & \multicolumn{4}{c}{MultiThumos} & \multicolumn{4}{c}{Charades}\tabularnewline
 & & & 0.2 & 0.5 & 0.7 & \textbf{Avg.} & 0.2 & 0.5 & 0.7 & \textbf{Avg.} \\

    \hline
    PDAN~\cite{dai2021pdan} & RGB & I3D & - & - & - & 17.3  & - & - & - & 8.5  \tabularnewline
    Coarse-Fine~\cite{kahatapitiya2021coarsefine} & RGB & I3D & - & - & -& - & - & - & - & 6.1  \tabularnewline
    MLAD~\cite{mlad} & RGB & I3D & - & - & - & 14.2  & - & - & - & - \tabularnewline
    MS-TCT~\cite{dai2022mstct} & RGB & I3D & - & - & - & 16.3  & - & - & - & 7.9   \tabularnewline
    PointTAD~\cite{tanpointtad} & RGB & I3D-E2E & 36.8 & 23.3 & 11.0 & 21.7 & 15.9 & 12.6 & 8.5 & 11.3 \tabularnewline
    PointTAD$\ddagger$~\cite{tanpointtad} & RGB & I3D-E2E & 39.7 & 24.9 & 12.0 & 23.5 & 17.5 & 13.5 & 9.1 & 12.1 \tabularnewline
   \hline
    ASL & RGB & I3D & \textbf{42.4} & \textbf{27.8} & \textbf{13.7} & \textbf{25.5} & \textbf{24.5} & \textbf{16.5} & \textbf{9.4} & \textbf{15.4} \tabularnewline  \bottomrule
  \end{tabular}
 }
\end{table*}

\begin{table*}[ht]
 \centering 
 \caption{\label{tab:result_ego4d} \textbf{Results on Ego4D-Moment Queries v1.0}. We report \textit{m}AP at different tIoU thresholds. Average \textit{m}AP in $[$0.1:0.1:0.5$]$ is reported on Ego4D-Moment Queries. Best results are in \textbf{bold}. EgoVLP, SF and OF denote EgoVLP~\cite{kevin2022egovlp}, Slowfast~\cite{slowfast} and Omnivore~\cite{girdhar2022omnivore} features. InterVideo~\cite{ego4dinternvideo} denotes features extracted from VideoMAE-L~\cite{tong2022videomae} and fine-tuned on Ego4D-Moment Queries.}
 \vspace{0.5em}
 \small
 {
  \begin{tabular}{l|l|cccc|c}
  \toprule
  \multirow{2}{*}{Method/Entry} & \multirow{2}{*}{Feature} & \multicolumn{4}{c}{mAP at IoUs, Val set} & \multicolumn{1}{c}{mAP at IoUs, Test set} \tabularnewline
 & & 0.1  & 0.3 & 0.5 & \textbf{Avg}. & \textbf{Avg}.   \tabularnewline
    \hline
    VSGN~\cite{vsgn} & SF & 9.10 & 5.76 & 3.41 & 6.03 & 5.68  \tabularnewline
    VSGN~\cite{kevin2022egovlp} & EgoVLP & 16.63 & 11.45 & 6.57 & 11.39 & 10.33  \tabularnewline
    ReLER~\cite{ego4dreler} & SF+OV & 22.75 & 17.61 & 13.43 & 17.94 & 17.67 \tabularnewline
    Actionformer~\cite{ego4dactionformer} & EgoVLP & 26.84 & 20.57 & 14.54 & 20.60 & - \tabularnewline
    Actionformer~\cite{ego4dactionformer} & EgoVLP+SF+OV & 28.26 & 21.88 & 16.28 & 22.09  & 21.76 \tabularnewline
    \textcolor{halfgray}{Actionformer~\cite{ego4dinternvideo}} & \textcolor{halfgray}{InternVideo} & \textcolor{halfgray}{-} & \textcolor{halfgray}{-} & \textcolor{halfgray}{-} & \textcolor{halfgray}{23.29}  & \textcolor{halfgray}{23.59}  \tabularnewline
   \hline
    ASL & EgoVLP & 29.45 & 23.03 & 16.08 & 22.83  & 22.25 \tabularnewline
    ASL & EgoVLP+SF+OV & \textbf{30.50} & \textbf{24.39} & \textbf{17.45} & \textbf{24.15} & \textbf{23.97}  \tabularnewline
 \bottomrule
  \end{tabular}
}
\end{table*}

\subsection{Datasets and Evaluation Metric}
\textbf{Datasets.}
To validate the efficacy of the proposed ASL, extensive experiments on 6 datasets of 3 types are conducted: i) densely-labeled: \textbf{MultiThumos}\cite{multithumos} and \textbf{Charades}\cite{charades}; ii) densely-labeled and egocentric: \textbf{Ego4D-Moment Queries v1.0}\cite{grauman2022ego4d} and \textbf{Epic-Kitchens 100}\cite{Damen2018EPICKITCHENS}; iii) single-labeled: \textbf{Thumos14}\cite{thumos} and \textbf{ActivityNet1.3}\cite{caba2015activitynet}. 
\par MultiThumos is a densely labeled dataset including 413 sports videos of 65 classes.  Charades is a large multi-label dataset containing 9848 videos of 157 action classes. These two datasets are both densely labeled and hence have multiple action instances in each video clip, where different actions may occur concurrently.
\par Ego4D-Moment Queries v1.0 (Ego4D-MQ1.0 for short) is a large-scale egocentric benchmark with 2,488 video clips and 22.2K action instances from 110 pre-defined action categories, which is densely labeled and composed of long clips. EPIC-Kitchens 100 is a large egocentric action dataset containing 100 hours of videos from 700 sessions capturing cooking activities in different kitchens. These two datasets are both large, egocentric and densely labeled.
\par Thumos14 is composed of 200 validation videos and 212 testing videos from 20 action classes while ActivityNet has 19,994 videos with 200 action classes. These two datasets are singly labeled and thus most of video clips in them have one action instance in each video clip.

 \textbf{Evaluation Metric.} 
Since ASL focuses on action detection, we take mean Average Precision (\textbf{mAP})
at certain tIoU thresholds as the evaluation metric. For all six datasets, we also report \textbf{average mAP} over several tIoU thresholds as the main metric.  The tIoU thresholds are set consistent with the official setup or previous methods, which is detailed in the caption of Table~\ref{tab:mthumosandcharades},~\ref{tab:result_ego4d},~\ref{tab:result_epickitchens},~\ref{tab:result_thumosandanet}.

\subsection{Implementation Details.} 
\par We follow the practice of using off-the-shelf pre-extracted features as input, specifically I3D~\cite{i3d} RGB features for MultiThumos, Charades, Thumos14 and ActivityNet , EgoVLP~\cite{kevin2022egovlp}, Slowfast~\cite{slowfast} and Omnivore~\cite{girdhar2022omnivore} features for Ego4D-MQ1.0, Slowfast features~\cite{slowfast,Damen2022RESCALING} for Epic-Kitchens 100.  
\par We train our model with a batch size of 2, 16, 2, 2 for 60, 30, 15, 25 epochs on MultiThumos, Charades, Ego4D-MQ1.0 and Epic-Kitchens 100 respectively, where the learning rate is set to $2e^{-4}$. On ActivityNet and Thumos, we train our model with the batch size of 16, 2, the learning rate of $1e^{-3}$, $1e^{-4}$  for 15, 30 epochs. We set $\lambda$ as 0.3 and $\theta$ as 0.2. 
\par In post-processing, we apply softNMS~\cite{softnms} to suppress redundant predictions. For fair comparison, we keep 200, 100, and 2000, 2000 predictions on Thumos14, ActivityNet, Ego4D-MQ1.0 and Epic-Kitchens 100 respectively. As on MultiThumos and Charades, considering that PointTAD~\cite{tanpointtad} splits a video clip into more than 4 parts and generates 48 predictions for each part, we keep 200 predictions on these two datasets. 
\par In the training process, we clamp $\sigma$ with a threshold (set as 5.0) to ensure $\sigma$ won’t be very large and thus prevent very small $p^{cls}, p^{loc}$, which may cause trivial solution to minimize the loss. Moreover, We tackle the issue of overlapped actions following~\cite{zhang2022actionformer,tian2019fcos}: i)use multi-scale mechanism~\cite{fpn} to assign actions with different
duration to different feature levels. ii)If a frame, even with multi-scale used, is still assigned to more than one ground-truth action, we choose the action with the shortest duration as its ground-truth target and model its action sensitivity based on this ground-truth.

\begin{table}[t]
    \centering
    \caption{\label{tab:result_epickitchens} \textbf{Results on EPIC-Kitchens 100 val set}. We report mAP at different tIoU thresholds and average mAP in $[0.1$:$0.1$:$0.5]$. All methods use the same SlowFast~\cite{slowfast,Damen2022RESCALING} features. }
    \vspace{0.5em}
    \small{
	\setlength{\tabcolsep}{2.5pt}
	\begin{tabular}{l|l|cccc}
        \toprule
	{Sub-Task} & {Method} & 0.1 &0.3 &  0.5 & Avg\\
		\hline
	\multirow{4}{*}{Verb} & BMN~\cite{lin2019bmn} & 10.8	&8.4 &5.6&8.4 \\
	& G-TAD~\cite{xu2020gtad} & 12.1 &  9.4 &  6.5 & 9.4\\
	& Actionformer~\cite{zhang2022actionformer} & 26.6 & 24.2  & 19.1 & 23.5 \\
        & ASL & \textbf{27.9} & \textbf{25.5} & \textbf{19.8} & \textbf{24.6} \\
	\hline
	\multirow{4}{*}{Noun} & BMN~\cite{lin2019bmn} & 10.3	&6.2&3.4&6.5 \\
	& G-TAD~\cite{xu2020gtad} & 11.0  & 8.6  & 5.4 & 8.4\\
	& Actionformer~\cite{zhang2022actionformer} & 25.2  & 22.7  & 17.0 & 21.9 \\
        & ASL & \textbf{26.0} & \textbf{23.4} & \textbf{17.7} & \textbf{22.6} \\
        \bottomrule
	\end{tabular}}
\end{table}

\begin{table*}[t]
 \centering 
 \caption{\label{tab:result_thumosandanet} \textbf{Results on Thumos14 and ActivityNet1.3}. We report \textit{m}AP at different tIoU thresholds. Average \textit{m}AP in $[$0.3:0.1:0.7$]$ is reported on THUMOS14 and $[$0.5:0.05:0.95$]$ on ActivityNet1.3. The best results are in \textbf{bold}. }
 \vspace{0.5em}
 \small
 {
  \begin{tabular}{l|l|cccccc|cccc} 
  \toprule
  \multirow{2}{*}{Model} & \multirow{2}{*}{Feature} & \multicolumn{6}{c}{Thumos14} & \multicolumn{4}{c}{ActivityNet1.3}\tabularnewline
 & & 0.3  & 0.4 & 0.5 & 0.6& 0.7 & \textbf{Avg.} & 0.5 & 0.75 & 0.95 & \textbf{Avg.} \\

    \hline
     BSN~\cite{BSN2018arXiv} & TSN~\cite{tsn} & 53.5 & 45.0 & 36.9 & 28.4 & 20.0 & 36.8 & 46.5 & 30.0 & 8.0 & 30.0 \tabularnewline 
    BMN~\cite{lin2019bmn} & TSN~\cite{tsn}& 56.0 & 47.4 & 38.8 & 29.7 & 20.5 & 38.5 & 50.1 & 34.8 & 8.3 & 33.9  \tabularnewline 
    G-TAD~\cite{xu2020gtad} & TSN~\cite{tsn} & 54.5 & 47.6 & 40.3 & 30.8 & 23.4 & 39.3 & 50.4 & 34.6 & 9.0 & 34.1\tabularnewline 
    P-GCN~\cite{PGCN2019ICCV} & I3D~\cite{i3d} & 63.6 &57.8 &49.1 & - & - & - & 48.3 & 33.2 & 3.3 & 31.1 \tabularnewline 
    TCANet~\cite{tcanet} & TSN~\cite{tsn} & 60.6 & 53.2 & 44.6 & 36.8 & 26.7 & 44.3 & 52.3 & 36.7 & 6.9 & 35.5 \tabularnewline 
    ContextLoc~\cite{contextloc} & I3D~\cite{i3d} & 68.3 & 63.8 & 54.3 & 41.8 & 26.2 & 50.9 & \textbf{56.0} & 35.2 & 3.6 & 34.2
   \tabularnewline 
    VSGN~\cite{vsgn} & TSN~\cite{tsn} & 66.7 & 60.4 & 52.4 & 41.0 & 30.4 & 50.2 & 52.4 & 36.0 & 8.4 & 35.1
     \tabularnewline 
     RTD-Net~\cite{rtdnet} & I3D~\cite{i3d} & 68.3 & 62.3 & 51.9 & 38.8 & 23.7 & 49.0 & 47.2  & 30.7 & 8.6 & 30.8
    \tabularnewline 
      SSN~\cite{SSN2017ICCV} & TS~\cite{ts} & 51.0 & 41.0 & 29.8 & - & - & - & 43.2 & 28.7 & 5.6 & 28.3 \tabularnewline 
      GTAN~\cite{gtan} & P3D~\cite{p3d} & 57.8 & 47.2 & 38.8 & - & - & -  & {52.6} & 34.1 & 8.9 & 34.3 \tabularnewline 
     AFSD~\cite{afsd} & I3D~\cite{i3d} & 67.3 & 62.4 & 55.5 & 43.7 & 31.1 & 52.0 & 52.4 & 35.3 & 6.5 & 34.4 \tabularnewline 
     React~\cite{shi2022react} & I3D~\cite{i3d} & 69.2 & 65.0 & 57.1 & 47.8 & 35.6 & 55.0 &49.6 & 33.0 & 8.6 & 32.6 \tabularnewline
     TadTR~\cite{tadtr} & I3D~\cite{i3d} & 62.4 & 57.4 & 49.2 & 37.8 & 26.3 & 46.6 & 49.1 & 32.6 & 8.5 & 32.3 \tabularnewline 
   Actionformer~\cite{zhang2022actionformer} & I3D~\cite{i3d} & 82.1 & 77.8 & 71.0 & 59.4 & 43.9 & 66.8 & 54.2 & 36.9 & 7.6 & 36.0 \tabularnewline
   \hline
    ASL & I3D~\cite{i3d} & \textbf{83.1} & \textbf{79.0} & \textbf{71.7} & \textbf{59.7} & \textbf{45.8} & \textbf{67.9} & 54.1 & \textbf{37.4} & 8.0 & \textbf{36.2} \tabularnewline
\bottomrule
\end{tabular}
 }
\end{table*}

\subsection{Main Results} 
\textbf{MultiThumos and Charades}: 
We compare ASL with state-of-the-art methods under detection-mAP on these two densely-labeled TAL benchmarks. PDAN\cite{dai2021pdan}, coarse-fine\cite{kahatapitiya2021coarsefine}, MLAD\cite{mlad}, MS-TCT\cite{dai2022mstct} are based on frame-level representation, while PointTAD\cite{tanpointtad} are query-based. As shown in Table~\ref{tab:mthumosandcharades}, ASL reaches the highest mAP over all tIoU thresholds, outperforming the previous best method(i.e. PointTAD) by 2.0\% absolute increase of average mAP on MultiThumos and 3.3\% on Charades. Notably, PointTAD is further trained in an end-to-end manner with strong image augmentation while ASL is feature-based, indicating that ASL performs more accurate TAL with more efficiency on densely-labeled datasets.

\textbf{Ego4D-MQ1.0 and Epic-Kitchens 100}:
These two datasets are both challenging as they are large-scale, egocentric, densely labeled and composed of longer clips. Table~\ref{tab:result_ego4d} reports the results on Ego4D-MQ1.0. The state-of-the-art methods are all based on Actionformer\cite{zhang2022actionformer} and perform frame-level recognition and localization with strong features. 
Using the same feature EgoVLP\cite{kevin2022egovlp}, ASL surpasses the current best entry\cite{ego4dactionformer}. Using the combined EgoVLP, slowfast\cite{slowfast} and omnivore\cite{girdhar2022omnivore} features, ASL gains 2.06\% improvement of average mAP on Val set and 2.21\% on Test set. Moreover, ASL performs better than~\cite{ego4dinternvideo} which uses a stronger but not open-sourced InternVideo~\cite{ego4dinternvideo} feature. 
Meanwhile, on Epic-Kitchens 100 as table~\ref{tab:result_epickitchens} shows, ASL outperforms the strong performance of Actionformer\cite{zhang2022actionformer}, BMN\cite{lin2019bmn} and G-TAD\cite{xu2020gtad} with the same Slowfast feature\cite{slowfast,Damen2022RESCALING}. The above results demonstrate the advantage of ASL on the challenging, egocentric and densely labeled benchmark. 

\textbf{Thumos14 and ActivityNet1.3}:
These two datasets are popular and nearly single-labeled, with approximately one action instance in each clip. Table~\ref{tab:result_thumosandanet} compares the results of ASL with various state-of-the-art methods (e.g., two-stage methods: BSN\cite{BSN2018arXiv}, G-TAD\cite{xu2020gtad}, P-GCN\cite{PGCN2019ICCV}, RTDNet\cite{rtdnet}, one-stage methods: AFSD\cite{afsd}, SSN\cite{SSN2017ICCV}, Actionformer\cite{zhang2022actionformer}.). On Thumos14, across all tIoU thresholds, ASL achieves the best and gains 1.1\% improvement of average mAP (67.9\% v.s. 66.8\%). On ActivityNet, ASL also outperforms previous methods of mAP@0.75 and average mAP, though the gap is slight. One possible reason is that due to the success of action recognition on ActivityNet, we follow the common practice~\cite{zhang2022actionformer,vsgn,contextloc} to fuse external video-level classification scores~\cite{cuhkanet}. In this case, class-level sensitivity will not play an important role in training. Another reason may be that since each video in ActivityNet is nearly single-labeled, our proposed ASCL will be short of positive and negative samples, leading to a non-significant increase compared to improvements on densely labeled datasets as Table~\ref{tab:mthumosandcharades},~\ref{tab:result_ego4d}.

\subsection{Ablation Study}

\label{sec:ablation} To further verify the efficacy of our contributions, we analyze main components of ASL on MultiThumos.

\begin{table}[t]
\caption{\label{tab:abl_component} \textbf{Ablation studies of components.} ASE: Action Sensitivity Evaluator. class.: class-level modeling. inst.: instance-level modeling. ASCL: Action Sensitive Contrastive Loss.}
\vspace{0.5em}
\small
{
\begin{tabular}{l|ccc|cccc}
\toprule
\multirow{4}{*}{\#} & \multicolumn{3}{c|}{Components} & \multicolumn{4}{c}{mAP at different tIoUs} \\
\hline
  & \multicolumn{2}{c}{ASE} &  ASCL & \multirow{2}{*}{0.2}  &  \multirow{2}{*}{0.5} & \multirow{2}{*}{0.7}  & \multirow{2}{*}{\textbf{Avg.}}   \\
 &  class. & inst. & & & & \\
 
\hline
 1& &  &                  & 39.6 &	25.9	& 11.6	& 23.4  \\
 2& \cmark & &            & 41.0 &	26.5 &	 12.9	& 24.5 \\ 
 3& & \cmark &            & 40.5     &   26.2    &    12.0      & 23.9 \\ 
 4&  &  & \cmark           & 40.2 & 26.1 & 11.8 & 23.7 \\
 5& \cmark & & \cmark     & 41.9 & 27.0 & 13.6 & 25.1 \\
 6& \cmark & \cmark&      & 41.8 & 27.2 &	13.3	& 25.0 \\ 
 \hline
 7 &\cmark & \cmark & \cmark & \textbf{42.4} &	\textbf{27.8}  &\textbf{13.7} & \textbf{25.5} \\
\bottomrule
\end{tabular}
}
\end{table}

\begin{table}[t]
\centering
\small
\caption{\label{tab:abl_gaussian}\textbf{Ablation studies of Gaussians weights.} cls and loc denotes classification and localization sub-task. For gaussian weights in class-level action sensitivity learning, \textbf{learnable}/\textbf{fixed} denotes parameters learnable/not learnable. \textbf{None} denotes not using gaussian weights.}
\vspace{0.5em}
\small
\begin{tabular}{c|c|c|ccccc}
    \toprule
	\# & {cls.} & {loc.} & 0.1 &0.3 &  0.5 & \textbf{Avg.}\\
        \hline
        1 & None & None & 40.9 & 26.3 & 12.3 & 24.2 \\ 
        \hline
        2 & \multirow{3}{*}{fixed} & None & 40.9 & 26.5 & 12.4 & 24.4 & \\
        3 &  & fixed & 41.0 & 26.6 & 12.7 & 24.6 \\
        4 & & learnable    & 41.7  & 26.8 & 13.0 & 24.9  \\
        \hline
        5 & \multirow{3}{*}{learnable} & None & 41.9 & 27.1 & 13.0 & 24.9 & \\
        6 &  & fixed & 42.0 & 26.9 & 13.4 & 25.1 \\
        7 & & learnable & \textbf{42.4} &\textbf{ 27.8} & \textbf{13.7} & \textbf{25.5} \\
    \bottomrule
\end{tabular}
\end{table}

 \textbf{Action Sensitive Evaluator.}  Our proposed ASE can be divided into class-level and instance-level modeling. we first investigate the effect of these parts. In Table~\ref{tab:abl_component}, baseline 1 denotes using our proposed framework without ASE and ASCL. After being equipped with class-level modeling, it boosts the performance by 1.1\% of average mAP (baseline 2 v.s. baseline 1). When further adding instance-level bias, it gains 0.5\% absolute increase (baseline 6 v.s. baseline 2). And our ASE contributes a total improvement of 1.6\% on average mAP (baseline 7 v.s. baseline 1). It is obvious that action sensitivity modeling from both class-level and instance-level is beneficial to TAL task.

\textbf{Gaussian Weights.} Then we analyze the effect of learnable gaussian weights in class-level action sensitivity learning. Table~\ref{tab:abl_gaussian} demonstrates that compared to baseline 1 which does not use any gaussian weights to learn action sensitivity, fixed gaussian weights with prior knowledge do bring benefits (baseline 2,3 v.s. baseline 1). Meanwhile, learnable gaussian weights are more favored (baseline 4 v.s. baseline 3, baseline 7 v.s. baseline 6). Moreover, learnable gaussian weights for both two sub-tasks achieve the best results. 
\par We further study the number of Gaussians used in classification and localization sub-task. As shown in Table~\ref{tab:abl_numgaussian}, using two Gaussians for localization and one Gaussian for classification achieves the best results. It is probably because on the one hand, using two Gaussians for localization explicitly allocates one for modeling start time and one for modeling end time. On the other hand, more Gaussian weights may be a burden for training, leading to inferior performance. 


\textbf{Action Sensitive Contrastive Loss.} Moreover, we delve into our proposed ASCL. As shown in Table~\ref{tab:abl_component}, ASCL improves around 0.6\% of average mAP on the basis of class-level prior (baseline 5 v.s. baseline 2) and 0.5\% on the basis of ASE (baseline 7 v.s. baseline 6). Baseline 4, where using ASCL alone denotes sampling near the center frame to form $f_{cls}$ and $f_{loc}$ directly, also gains an improvement of 0.3\% compared to the vanilla framework (baseline 4 v.s. baseline 1). This indicates the effectiveness of contrast between actions and backgrounds. When performing ASCL based on ASE, it will facilitate the final performance more because it can alleviate the misalignment as discussed in~\ref{sec:ascl}.

\par
\begin{figure}[t]
\centering
\subfigure[Ablation of $\lambda$]{
    \begin{minipage}{0.472\linewidth}
    \includegraphics[height=0.5825\linewidth, width=1.01\linewidth]{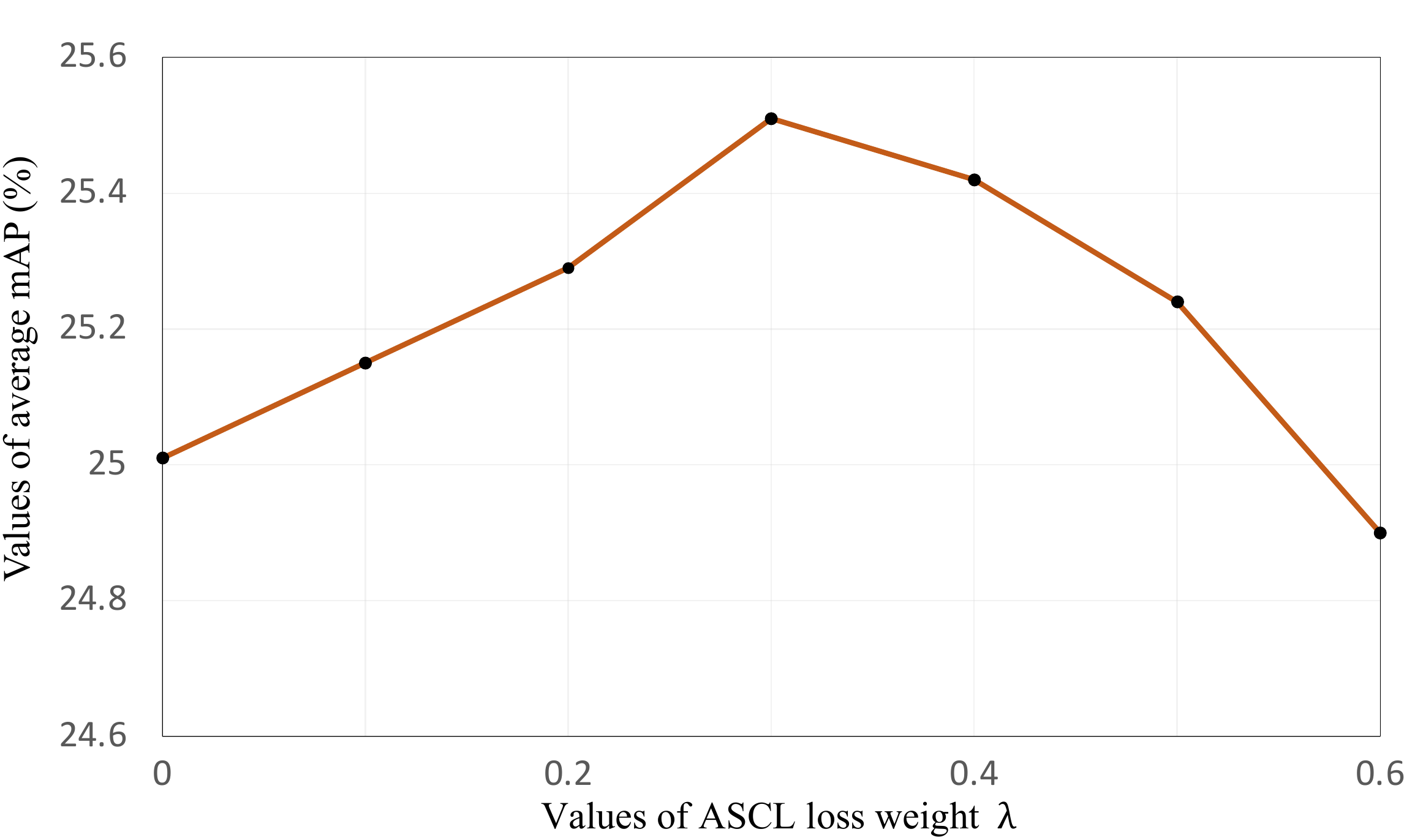}
    \end{minipage}
}
\subfigure[Ablation of $\delta$]{
    \begin{minipage}{0.472\linewidth}
    \includegraphics[height=0.6026\linewidth, width=1.05\linewidth]{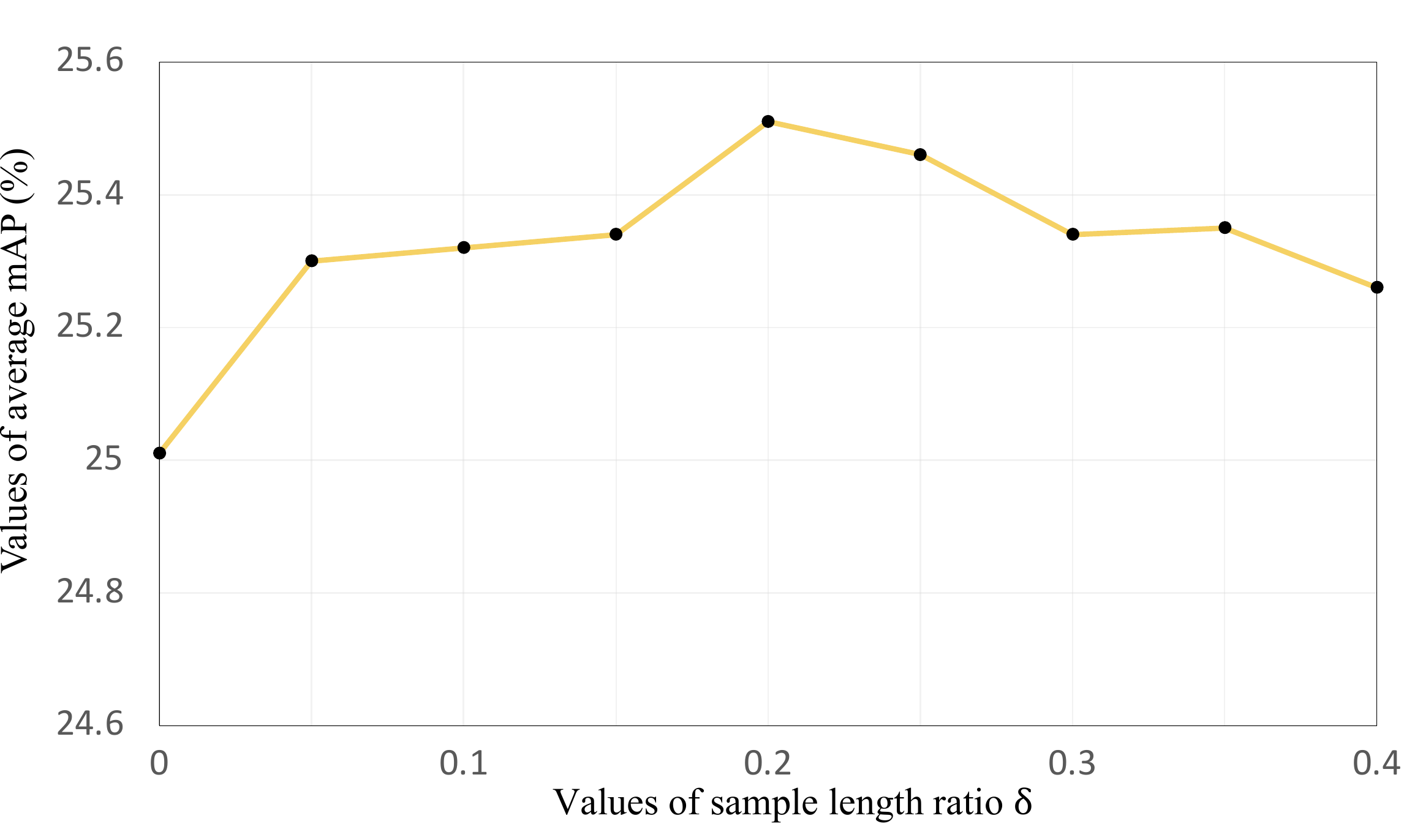}
    \end{minipage}
}
\caption{Ablation of hyperparameters in ASCL.}
\label{fig:abl_lambda}
\end{figure}

 Finally we discussed the hyperparameters in ASCL. Fig~\ref{fig:abl_lambda}(a) shows the performance curve of average mAP corresponding to ASCL weight $\lambda$. Average mAP on MultiThumos generally improves when $\lambda$ increases and slightly drop as $\lambda$ reaches 0.4. Fig~\ref{fig:abl_lambda}(b) reports the average mAP to different sampling length ratios $\delta$. When $\delta$ equals 0.2, our method achieves the best. In this case, we set $\lambda$ to 0.3 and $\delta$ to 0.2. 

\begin{table}[t]
\centering
\small
\caption{\label{tab:abl_numgaussian}\textbf{Ablation studies of number of Gaussians weights.} \#cls and \#loc denote the number of Gaussian weights used in classification and localization sub-task. \textbf{shared} indicates two sub-tasks share one Gaussian weights.}
\vspace{0.5em}
\small
\begin{tabular}{c|c|ccccc}
        \toprule
	{\#cls} & {\#loc} & 0.1 &0.3 &  0.5 & Avg\\
		\hline
        1(shared) & 1(shared)     & 42.2 & 27.2 & \textbf{13.7} &25.3 \\ 
        \hline
	\multirow{3}{*}{0} & 0     & 40.9 & 26.3 & 12.3  &  24.2      \\
	& 1                        & 41.5 &  26.9 & 13.0 & 24.8\\
	& 2                        & 41.6 &  27.1  & 13.4 & 25.0 \\
	\hline
	\multirow{3}{*}{1} & 0     & 42.2 & 27.1 & 13.2 & 25.1 \\
	& 1                        & 42.0 & 26.7 & 13.1 & 24.9 \\
	& 2                        & \textbf{42.4} & \textbf{27.8} & \textbf{13.7} &\textbf{25.5} \\
        \hline
        \multirow{3}{*}{2} & 0     & 42.3 & 26.9 & 13.3 &    25.1 \\
	& 1                        & 41.8 & 26.9 & 13.0 &    25.0 \\
	& 2                        & 42.0 & 27.2 & 13.6 &   25.3 \\
        \bottomrule
\end{tabular}
\vspace{-0.5em}
\end{table}

\begin{figure}[t]
\centering
\includegraphics[height=1.130\linewidth, width=1.02\linewidth]{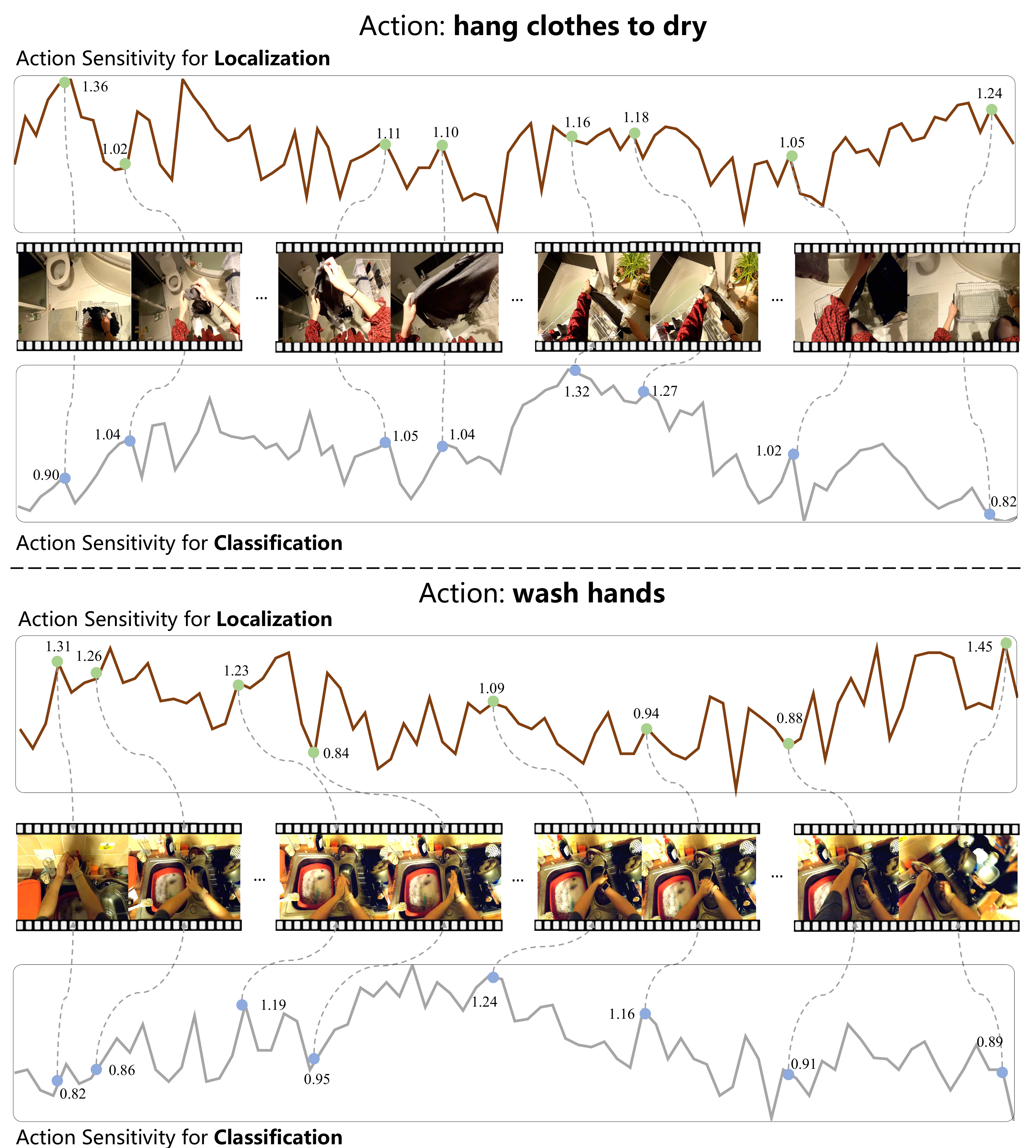}
\caption{Visualization of (\textbf{Top}) the frame sensitivity to sub-tasks of Action: \textbf{hang clothes to dry} and (\textbf{bottom}) Action: \textbf{wash hands}. Please zoom in for the best view.}
\label{fig:visualize}
\end{figure}

\subsection{Qualitative Experiment}
To better illustrate the effectiveness of ASL, we visualize some qualitative results of Ego4D-MQ1.0 benchmark in Fig~\ref{fig:visualize}. We show that i) frames depicting action's main sub-action (i.e., \textit{hang clothes on the hanger}, \textit{water run through hands}) are of higher action sensitivity for classification. ii) Frames depicting near-start and near-end sub-action (i.e, turn the tap on, lift laundry basket, e.t.c.) are of higher action sensitivity for localization. Moreover, action sensitivity of frames is not continuous, as our proposed instance-level action sensitivity is discrete partly because blurred or transitional frames exist in video clips.

\vspace{2.3em}
\section{Conclusion}
In this paper, we introduce an Action Sensitivity Learning framework (ASL) for temporal action localization (TAL). ASL models action sensitivity of each frame and dynamically change their weights in training. Together with the proposed Action Sensitive Contrastive Loss (ASCL) to further enhance features and alleviate misalignment,  ASL is able to recognize and localize action instances effectively.  For accurate TAL, fine-grained information should be considered (e.g. frame-level information). We believe that ASL is a step further in this direction. In the future, efforts could be paid to more complicated sensitivity modeling. Besides, ASL could also be redesigned as a plug-and-play component that will be beneficial to various TAL methods.

\textbf{Acknowledgements:} This work is supported by the Fundamental Research
Funds for the Central Universities (No.226-2023-00048) and Major Program of the National Natural Science Foundation of China (T2293720/T2293723)

\clearpage
{\small
\bibliographystyle{ieee_fullname}
\bibliography{egbib}
}

\end{document}